\UseRawInputEncoding
\documentclass{article}
\PassOptionsToPackage{square, numbers}{natbib} 
\PassOptionsToPackage{pdftex}{graphicx}

\usepackage[preprint]{neurips_2024}

\usepackage[dvipsnames]{xcolor}

\usepackage[utf8]{inputenc} 
\usepackage[T1]{fontenc}    
\usepackage{hyperref}       
\usepackage{url}            
\usepackage{booktabs}       
\usepackage{amsfonts}       
\usepackage{nicefrac}       
\usepackage{microtype}      
\usepackage{xcolor}         
\usepackage[ruled,vlined]{algorithm2e}
\usepackage{graphicx}
\usepackage{soul}
\usepackage{tikz}
\usepackage{amssymb}
\usepackage[justification=centering]{caption}

\usepackage{wrapfig}
\usepackage{caption}
\usepackage{mathtools}
\usepackage{multirow}
\usepackage{algpseudocode}
\usepackage{comment}
\usepackage{subcaption}
\usepackage{listings}
\usepackage{float}
\usepackage{enumitem}
\usepackage{listings}
\usepackage{xcolor}
\usepackage[pdftex]{graphicx}

\setlist[itemize]{
  nosep,
  align=left,
  leftmargin=0pt,
  labelwidth=1.25em,
  itemindent=1.25em,
  labelsep=0pt,
}
\setlist*[itemize,2]{
  leftmargin=1.25em,
}
\usepackage{xspace}
\usepackage{enumitem}
\newcommand{\genPipeline}{{\texttt{Pruned MCTS}}\xspace}
\newcommand{\shortname}{{\texttt{SPHERE}}\xspace}

\newcommand{\qwenOne}{{\text{Qwen2.5-1.5B}}\xspace}

\newcommand{\qwenSeven}{{\text{Qwen2.5-7B}}\xspace}
\newcommand{\qwenSevenMath}{{\text{Qwen2.5-7B-Math}}\xspace}
\newcommand{\PHI}{{\texttt{phi-4}}\xspace}

\newcommand{\qwenThreeFull}{{\text{Qwen/Qwen2.5-3B-Instruct}}\xspace}
\newcommand{\qwenSevenFull}{{\text{Qwen/Qwen2.5-7B-Instruct}}\xspace}

\title{Self-Evolved Preference Optimization for Enhancing Mathematical Reasoning in Small Language Models}

\author{
  Joykirat Singh$^{1}$, 
  Tanmoy Chakraborty$^{2}$, 
  Akshay Nambi$^{1}$ \\
  $^{1}$Microsoft Research, India \\
  $^{2}$IIT Delhi, India \\
  \texttt{tanchak@ee.iitd.ac.in, akshayn@microsoft.com}
}

\begin{document}
\maketitle

\begin{abstract}
Large language models (LLMs) have significantly improved their reasoning capabilities; however, they still struggle with complex multi-step mathematical problem-solving due to error propagation, lack of self-correction, and limited adaptability to diverse reasoning styles. Existing methods rely on static fine-tuning or prompt engineering, which fail to generalize across problem complexities, while the scarcity of high-quality preference data further hinders reliable reasoning.

We introduce \shortname, a self-evolving data generation pipeline that enhances reasoning in small language models (SLMs) by iteratively generating, correcting, and diversifying reasoning chains. \shortname\ operates in three stages: (i) Self-Generation, where the model autonomously constructs problem-solving steps; (ii) Self-Correction, enabling it to identify and rectify errors; and (iii) Diversity Induction, improving robustness through multiple valid reasoning trajectories. This self-evolution mechanism strengthens mathematical reasoning and enhances model reliability. Evaluations on MATH 500, GSM8K, AIME, AMC, and Olympiad show that \shortname-trained models achieve significant gains over their base versions and match/surpass GPT-4o on certain benchmarks. Our findings demonstrate that self-evolving models can close the reasoning gap between SLMs and state-of-the-art LLMs, making mathematical AI more reliable, scalable, and efficient.

\end{abstract}
\section{Introduction}
Despite advancements in LLMs, complex multi-step reasoning remains a challenge. While scaling improves general understanding, structured tasks like mathematical problem-solving and code generation require logically consistent steps, error correction, and iterative refinement. Post-training techniques have shown promise in enhancing reasoning efficiency beyond pre-training by aligning models with task-specific objectives~\cite{ouyang2022traininglanguagemodelsfollow}. However, reward-based optimization and reinforcement learning struggle with generating high-quality supervision signals, as error propagation can degrade multi-step reasoning performance.

Recent post-training methods refine reasoning by evaluating intermediate steps rather than final answers. Process-based reward models~\cite{uesato2022solvingmathwordproblems, lightman2023letsverifystepstep} offer finer-grained supervision but rely on handcrafted reward functions that lack generalization. Reinforcement learning~\cite{kumar2024traininglanguagemodelsselfcorrect} iteratively refines reasoning traces but suffers from reward sparsity and policy collapse in long-horizon tasks. Search-based methods like Monte Carlo Tree Search (MCTS)~\cite{xie2024montecarlotreesearch} and Beam Search~\cite{feng2024alphazeroliketreesearchguidelarge, trinh2024solving, xin2024deepseekproverv15harnessingproofassistant} improve step-wise reasoning by exploring multiple paths; yet they are computationally expensive and prone to biased exploration, limiting the diversity of reasoning trajectories.

A key challenge in multi-step reasoning is generating supervision signals that balance correctness, diversity, structured exploration. Existing methods reinforce correct solutions but overlook structured errors -- arithmetic mistakes, logical flaws, or poor decompositions -- that could aid learning. Rejection sampling filters low-confidence outputs but lacks active guidance for better reasoning~\cite{luo2024improvemathematicalreasoninglanguage}.

Direct Preference Optimization (DPO)~\cite{rafailov2024directpreferenceoptimizationlanguage} optimizes models using preference-ranked data instead of explicit rewards, but its effectiveness in multi-step reasoning is limited by the lack of high-quality preference data capturing both correct and incorrect reasoning~\cite{chen2024steplevelvaluepreferenceoptimization, lu2024stepcontrolleddpoleveragingstepwise}. Mathematical reasoning requires structured evaluation beyond fluency, yet human-annotated datasets focus on correctness, overlooking the value of structured errors~\cite{lai2024stepdpostepwisepreferenceoptimization}. Manual curation is also costly, expertise-intensive, and impractical at scale, restricting its utility.

\begin{figure}[t]
\centering
  \includegraphics[width=0.65\textwidth]{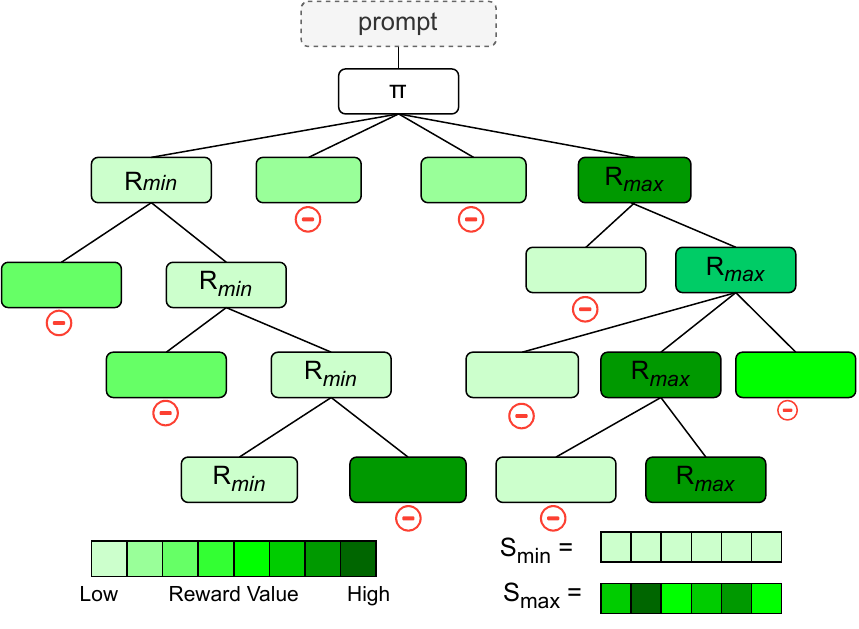}
  \caption{Illustration of Pruned MCTS Rollouts.}
  \label{fig:mcts-pruned}
  \vspace{-10pt}
\end{figure}

To address these challenges, we introduce \shortname\ -- Self-Evolved Preference Optimization for Reasoning, a fully automated pipeline for generating preference data without human annotation. Using MCTS rollouts, our method identifies high-reward (correct) and low-reward (flawed) reasoning paths, $S_{max}$ and $S_{min}$, respectively. Unlike prior approaches that discard suboptimal solutions \cite{zhao2024marcoo1openreasoningmodels, xie2024montecarlotreesearch}, \shortname\ refines both correct and incorrect reasoning through structured self-correction, enriching training with diverse preference pairs. By integrating these reasoning traces into DPO, \shortname\ eliminates reliance on costly human annotations while addressing biased search, reward sparsity, and data scarcity, enhancing model robustness in multi-step reasoning.

While MCTS is widely used in structured reasoning, applying it to preference data generation presents unique challenges. Expanding the full reasoning tree is computationally infeasible due to exponential search space growth. To address this, we propose a \textit{pruned MCTS rollout strategy} that leverages process-based reward models~\cite{lightman2023letsverifystepstep} for fine-grained step-wise evaluation. At each decision level, we retain only the highest-reward ($S_{max}$) and lowest-reward ($S_{min}$) reasoning paths, pruning all other branches. This targeted selection captures the most informative trajectories while significantly reducing computational cost (see Figure~\ref{fig:mcts-pruned}). By focusing on both optimal and flawed paths, we efficiently curate high-quality preference data for DPO without exhaustive tree expansion. We refine data quality and diversity through a three-stage self-evolution pipeline.

\begin{figure*}[t]
  \includegraphics[width=\textwidth]{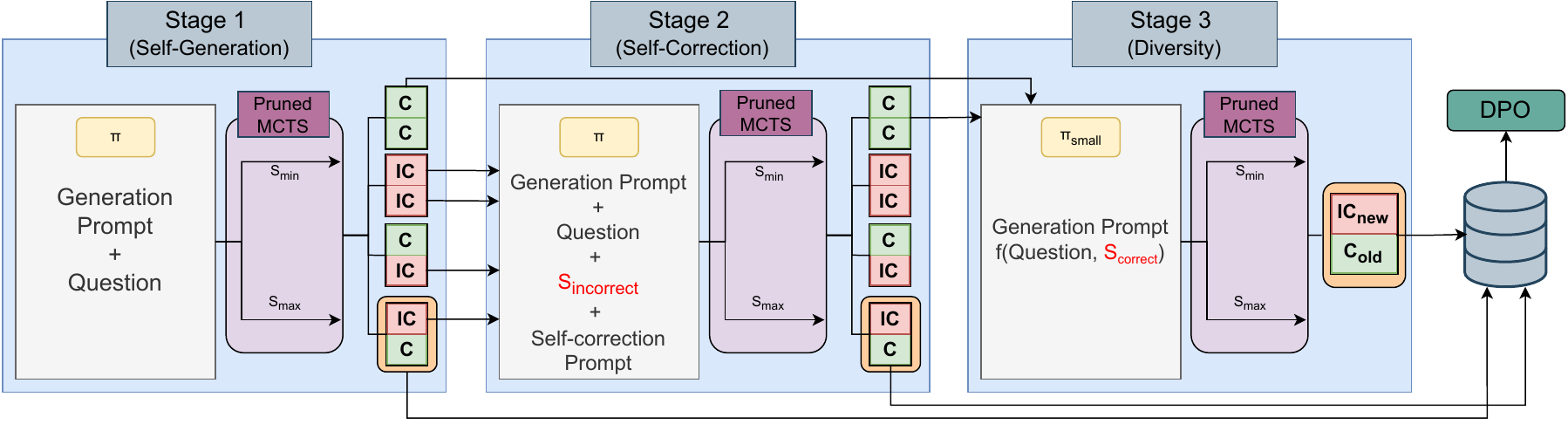}
  \caption{Illustration of all stages in \genPipeline. \textcolor{green}{C} and \textcolor{red}{IC} denote $Sol_{max}$ (correct solution) and $Sol_{min}$ (incorrect solution) extracted from each rollout. Reasoning pairs within the \textcolor{Goldenrod}{Gold Box} are selected for preference learning.}
  \label{fig:mcts-pruned}
  \vspace{-5pt}
\end{figure*}

 \textbf{Stage 1: Self-Generation }– The model generates reasoning paths, but not all rollouts produce valid contrastive preference pairs ($S_{max},S_{min}$). Some cases lack a meaningful preference contrast, requiring further refinement.

 \textbf{Stage 2: Self-Correction} –  The model self-reflects on errors, revising incorrect paths to ensure that $S_{max}$ represents a valid solution and $S_{min}$ retains contrastive value. This process improves the reliability of preference data.

 \textbf{Stage 3: Diversity Generation} – To capture varied reasoning mistakes, a smaller model regenerates $S_{min}$, introducing controlled errors that enhance failure diversity and generalization.

Our pipeline systematically generates diverse, self-improving preference data by integrating self-generation, self-correction, and diversity. Training SLMs via DPO on this data enables them to surpass their base versions and even outperform larger models like GPT-4o. By learning from both optimal and flawed reasoning, \shortname\ surpasses standard preference-based fine-tuning, highlighting the power of self-evolution in multi-step reasoning.
Our key contributions are as follows\footnote{We have attached the source code and we are committed to releasing them after accepting of the paper.}: 
\begin{itemize}
[noitemsep,nolistsep,topsep=0pt,leftmargin=1em]
    \item \textbf{Pruned MCTS for Efficient Preference Data:}  A reward-guided pruning strategy selects only the highest- and lowest-reward reasoning paths, reducing computation while ensuring high-quality contrastive data for DPO.
 \item \textbf{Three-stage Self-Evolution for Data Refinement:} A fully automated pipeline with (i) self-generation (initial reasoning paths), (ii) self-correction (model-driven error refinement), and (iii) diversity generation (error augmentation). 
 \item{\bf High-Quality Step-wise Preference Data:} Fine-grained preference captures both correct and incorrect reasoning, helping models learn from optimal solutions and systematic mistakes.
 \item{\bf Superior Multi-Step Reasoning Performance:} SLMs, fine-tuned with \shortname, outperform their base versions and surpass GPT-4o in math problem-solving and even improving performance of models like DeepSeek-R1-Distill-Qwen-7B~\cite{deepseekai2025deepseekr1incentivizingreasoningcapability} by an average of 5.1\%, demonstrating a scalable, annotation-free approach for reasoning.
 \end{itemize}

Our self-evolution pipeline improves RL frameworks and pre-training by addressing data quality constraints. In methods like DeepSeek~\cite{deepseekai2025deepseekr1incentivizingreasoningcapability}, it generates diverse, high-preference reasoning trajectories, enhancing training. Integrated with PRMs and RL, it helps models internalize structured reasoning, improving generalization. By curating accurate, self-corrected, and diverse preference data, it offers a scalable, annotation-free solution for multi-step reasoning in LLMs.

\section{Related Work}
Recent LLM performance relies heavily on high-quality data curation. Early methods like MAmmoTH~\cite{yue2023mammothbuildingmathgeneralist} and Open-MathInstruct~\cite{toshniwal2024openmathinstruct118millionmath} use synthetic data from stronger teachers and supervised fine-tuning (SFT) to boost accuracy on benchmarks like GSM8K~\cite{cobbe2021trainingverifierssolvemath} and MATH~\cite{hendrycks2021measuringmathematicalproblemsolving}. However, these approaches confine models to their teacher’s reasoning, discarding problems beyond the teacher’s grasp.
Recently, process reward models (PRMs)~\cite{uesato2022solvingmathwordproblems, lightman2023letsverifystepstep} have been introduced, which provide feedback to each step of the Chain-of-thought reasoning. The feedback from PRMs coupled with algorithms such as PPO~\cite{schulman2017proximalpolicyoptimizationalgorithms} and DPO~\cite{rafailov2024directpreferenceoptimizationlanguage} have been used to further enhance the model's capabilities. For example, ~\citet{xie2024montecarlotreesearch, chen2024steplevelvaluepreferenceoptimization}  proposed to use MCTS to collect preference data, and employed DPO to update the LLM policy using the step level preference data. ~\citet{guan2025rstarmathsmallllmsmaster} also leveraged MCTS to generated huge amount of synthetic data, and through multiple rounds of self evolution, fine-tuned an SLM and a PRM to build a System 2 model.~\citet{zhao2024marcoo1openreasoningmodels} used multiple MCTS rollouts as an SFT dataset, while ~\citet{chen2024alphamathzeroprocesssupervision} integrated a value model with the LLM to generate process supervision and step-level evaluation signals in MCTS.

Unlike exhaustive MCTS rollouts, our approach employs reward-guided MCTS pruning to select high-contrast reasoning paths while lowering computational costs. We also introduce a structured self-evolution pipeline to generate diverse, high-preference reasoning trajectories, enhancing mathematical reasoning and self-correction in LLMs.
\section{Proposed Methodology}


\shortname is a self-evolution framework that enhances multi-step reasoning in SLMs by generating high-quality preference data without human supervision. It leverages MCTS to explore reasoning trajectories efficiently while using a process-based reward model to assign step-wise correctness scores. To mitigate computational costs, \shortname prunes suboptimal branches, retaining only the highest-reward ($S_{max}$) and lowest-reward ($S_{min}$) paths per rollout. This selective sampling produces high-quality preference pairs, enabling models to learn both optimal strategies and systematic failure patterns. By focusing on the most informative reasoning trajectories, \shortname ensures scalable, efficient preference data generation for training SLMs without annotation.
\shortname employs a three-stage self-evolution process:

\subsection{Stage 1: Self-Generation of Reasoning Trajectories}

The first stage of \shortname constructs structured reasoning trajectories by using a base SLM to explore diverse problem-solving paths. Given a policy $\pi$ and a dataset $\mathcal{D}$ with question-answer pairs, $\pi$ generates multi-step reasoning sequences at a high temperature to enhance variability. At each time step $t$, the model generates $E$ distinct reasoning steps:
\begin{equation}
\small
S_t = \{ S_t^1, S_t^2, \dots, S_t^E \}, \quad S_t^i \sim \pi(\cdot | S_{t-1})
\end{equation}
where $S_t^i$ represents the $i$-th candidate reasoning step at step $t$.

To ensure efficient exploration, only two steps per rollout are retained: (1) $S_t^{\max}$: The step \textbf{most likely} to lead to the correct final answer. (2) $S_t^{\min}$: The step \textbf{least likely} to lead to the correct final answer but still plausible.

These steps are scored using PRM, $\pi_{\text{prm}}$, which evaluates their likelihood of leading to the correct solution:
\begin{equation}
\small
S_t^{\max} = \max_{S_t^i} R(S_t), \quad S_t^{\min} = \min_{S_t^i} R(S_t)
\end{equation}
where $R(S_t)$ assigns a reward score (detailed in Section \ref{sec:ra}). The process continues recursively until reaching a final answer or a predefined depth limit, forming two complete reasoning trajectories: (1) $Sol_{\max}$: A sequence of steps composed of $S_t^{\max}$, forming the most optimal reasoning trajectory. (2) $Sol_{\min}$: A sequence of steps composed of $S_t^{\min}$, forming the weakest but still structured reasoning trajectory.

\subsubsection{Reward Assignment for Reasoning Steps}\label{sec:ra}

The reward function in \shortname is designed to evaluate intermediate reasoning steps, ensuring that each step is assigned a structured preference score. A \textbf{Process Reward Model (PRM)} $\pi_{\text{prm}}$ is used to assign scores between \([0,1]\), where
$1$ indicates a high likelihood of leading to the correct final answer, and $0$ indicates a highly unreliable reasoning step. For an initial step $s_0$, the reward is directly assigned as: $
R(s_0) = \pi_{\text{prm}}(s_0). $
For subsequent steps $s_t$, we incorporate an advantage reward that accounts for progress made:
\begin{equation}
\small
R(s_t) = \pi_{\text{prm}}(s_t) + \frac{\pi(s_t) - \pi(s_{t-1})}{\pi_{\text{prm}}(s_{t-1})}
\end{equation}
This additional term ensures that intermediate reasoning steps are not only individually evaluated but also assessed based on their ability to improve upon prior steps, creating a progressive refinement signal for training.

\subsubsection{Handling Missing $S_{\min}$ and $S_{\max}$ Cases}
In some cases, both $S_{\max}$ and $S_{\min}$ may be missing due to:
\textbf{1. All solutions being incorrect}: The model fails to produce any invalid reasoning paths, preventing the identification of a meaningful $S_{\min}$. \textbf{2. All solutions being correct}: The generated reasoning steps exhibit only valid problem-solving approaches, leading to a lack of contrastive training pairs. To address the first and second gaps, Stage 2: Self-Correction and Stage 3: Diversity are introduced, respectively.

\subsection{Stage 2: Self-Correction for Preference Data Generation}

In this stage, we enhance the model’s self-correction capability by prompting it to reflect on its own reasoning, identify mistakes, and regenerate improved solutions. The self-correction dataset is specifically constructed from cases where both $Sol_{\max}$ or $Sol_{\min}$ result in an incorrect final answer, meaning the model initially fails to produce a valid reasoning trajectory.

To generate preference pairs for self-correction, we apply the same MCTS-guided exploration approach used in Stage 1. This process yields:
\begin{itemize}
    \item \( Sol^{SC}_{\max} \): The self-corrected solution with the highest probability of reaching the correct final answer.
    \item \( Sol^{SC}_{\min} \): The self-corrected solution with the lowest probability of reaching the correct final answer but still plausible.
\end{itemize}

These self-corrected reasoning paths allow the model to iteratively refine its problem-solving ability, ensuring that it learns from its own structured errors rather than discarding them.

\subsubsection{Handling Incorrect Initial Reasoning}

Given that the dataset for this phase consists of cases where the original model outputs are incorrect, the self-correction process must effectively generate contrastive solutions. This is achieved by, \textbf{(i)} prompting the model to critically evaluate each step of its reasoning and identify where it deviated from correctness; \textbf{(ii)} introducing an additional shaping term in reward function, to ensure \( Sol^{SC}_{\max} \) is superior to previous incorrect solution.

By structuring self-corrected preference data in this manner, we enable the model to develop a more refined understanding of failure cases and improve its overall reasoning capability (see Appendix~\ref{app:prompts} for the prompts used).

\subsubsection{Reward Assignment for Self-Correction}

To ensure that the self-corrected solution \( Sol^{SC}_{\max} \) is superior to the previous incorrect solutions (\( Sol_{\max} \) or \( Sol_{\min} \)), we introduce an additional shaping term in the reward function. This shaping term is derived from the \textbf{Outcome-Supervised Reward Model (ORM)}~\cite{uesato2022solvingmathwordproblems}, which differs from PRMs by evaluating complete reasoning trajectories instead of individual steps. The modified reward function for self-corrected reasoning pairs is defined as:
\[
\small
R(s_t) =
\begin{cases}
    \pi_{\text{prm}}(s_t) + \text{ORM}(Sol), & t = 0 \\\\
    \pi_{\text{prm}}(s_t) + \text{ORM}(Sol) + \\\frac{\pi(s_t) - \pi(s_{t-1})}{\pi_{\text{prm}}(s_{t-1})}, & t > 0
\end{cases}
\]
where \( \text{ORM}(Sol) \) provides a scalar score for the entire reasoning trajectory rather than individual steps. \( \pi_{\text{prm}}(s_t) \) assesses step-wise reasoning quality. The advantage term ensures that corrections contribute progressively to a better solution. With ORM-based outcome supervision, \shortname ensures that self-correction fixes errors as well as improves generalization from flawed reasoning.

\subsection{Stage 3: Enhancing Diversity in Preference Data}

While the previous two stages effectively generate high-quality preference data, we observe that in 68\% cases, both model-generated solutions $Sol_{\max}$ and $Sol_{\min}$ lead to the correct final answer. This results in a lack of contrastive negative examples, which are crucial for robust preference learning. To address this, we introduce a diversity enhancement mechanism that strategically introduces incorrect reasoning samples while maintaining preference-based supervision.

\subsubsection{Generating Diverse Incorrect Reasoning}
To introduce more diversity, we utilize a smaller model $\pi_{\text{small}}$, which shares the same architecture as the original policy $\pi$ but has fewer parameters. This smaller model explores alternative reasoning paths with a higher likelihood of generating incorrect yet plausible solutions. The diversity augmentation process is structured as follows:\\
\textbf{1. Targeting Overlapping Correct Solutions}: We identify instances where both $Sol_{\max}$ and $Sol_{\min}$ in previous stages resulted in correct final answers.\\
\textbf{2. Wider Exploration with $\pi_{\text{small}}$}: The smaller model $\pi_{\text{small}}$ is tasked with generating reasoning trajectories for these cases, using an expanded exploration budget of $2E$ to increase the probability of producing diverse errors.\\
\textbf{3. Filtering via MCTS-Guided Selection}: The same MCTS mechanism is applied to extract the most and least promising reasoning steps, ensuring structured error diversity.

By leveraging a smaller model for targeted incorrect reasoning generation, we introduce meaningful contrastive pairs that enrich the dataset and improve model robustness in handling diverse failure cases.

\subsubsection{Reward Assignment for Diversity Enhancement}

The reward framework remains consistent with the previous stages. Specifically,
(i) the PRM continues to assess intermediate reasoning steps and (ii) the ORM ensures that incorrect solutions generated by $\pi_{\text{small}}$ remain structurally valid yet distinct from existing correct solutions.

Through this three-stage self-evolved data curation pipeline, \shortname\ ensures a comprehensive balance between correctness, diversity, and structured exploration, leading to improved multi-step reasoning capabilities in SLMs. See Appendix~\ref{app:mcts_example} for illustrative examples for self evolved reasoning chains.

\subsection{Preference Tuning via DPO}
We fine-tune the policy model using DPO \cite{rafailov2024directpreferenceoptimizationlanguage}, which efficiently aligns model reasoning without explicit reward modeling, ensuring scalability.

Given an input prompt \(x\) and a preference pair \((y_{\text{c}}, y_{\text{r}})\), where \(y_{\text{c}}\) is the preferred (higher-quality) reasoning trajectory and \(y_{\text{r}}\) is the less desirable one, DPO maximizes the likelihood of selecting \(y_{\text{c}}\) while minimizing that of \(y_{\text{r}}\). The optimization objective is defined as:

\begingroup
\small
\begin{align}
    \mathcal{L}_{DPO}(\theta) = 
    & -\mathbb{E}_{(x, y_{\text{c}}, y_{\text{r}}) \sim D} 
    \Bigg[
    \log \sigma \Bigg(
    \beta \log \frac{\pi_{\theta}(y_{\text{c}} \mid x)}
    {\pi_{\text{ref}}(y_{\text{c}} \mid x)} - \beta \log \frac{\pi_{\theta}(y_{\text{r}} \mid x)}
    {\pi_{\text{ref}}(y_{\text{r}} \mid x)}
    \Bigg)
    \Bigg]
\end{align}
\endgroup
where \(D\) represents the set of preference pairs, \(\sigma\) is the sigmoid function, \(\pi_\theta(\cdot \mid x)\) is the policy model being optimized, and \(\pi_{\text{ref}}(\cdot \mid x)\) is the reference model, which remains unchanged during training. The hyperparameter \(\beta\) controls the divergence from the reference model, ensuring stability in learning.

The preference dataset used in the DPO training consists of step-wise and trajectory-level comparisons from the three-stage self-evolution process: \textbf{1. Self-Generated Preferences}: Pairs \( (Sol_{\max}, Sol_{\min}) \) capturing high-confidence and low-confidence reasoning paths. \textbf{2. Self-Corrected Preferences}: Pairs \( (Sol_{\max}^{SC}, Sol_{\min}^{SC}) \) where the model refines its own incorrect reasoning. \textbf{3. Diversity-Enhanced Preferences}: Pairs extracted from a smaller model \(\pi_{\text{small}}\), enriching the dataset with structured incorrect reasoning.

By integrating these preference pairs, the model learns to differentiate between well-structured and flawed reasoning while improving its ability to generalize across diverse problem-solving patterns.

\section{Experimental Setup \& Implementation}
We evaluate the effectiveness of \shortname\ in generating high-quality self-evolved preference datasets and improving mathematical reasoning in SLMs.

\subsection{Model Architectures \& Training Setup}
\paragraph{Dataset Generation Models.} For dataset generation and preference training, we employed multiple models of varying sizes and architectures to balance efficiency and reasoning diversity:
\begin{itemize}
\item Generation Policy ($\pi$): \qwenSevenFull~\cite{qwen2025qwen25technicalreport}, responsible for generating multi-step reasoning trajectories.
\item Diversity Augmentation Model ($\pi_{small}$): \qwenThreeFull \cite{qwen2025qwen25technicalreport}, used to enhance diversity by generating alternative reasoning steps and incorrect solutions.
\item Process Reward Model (PRM, $\pi_{prm}$): Qwen/Qwen2.5-Math-PRM-7B~\cite{prmlessons}, trained to assess the quality of intermediate reasoning steps.
\end{itemize}

During reasoning trajectory generation, the base policy $\pi$ generates 5 reasoning steps per prompt at a sampling temperature of 0.8. $\pi_{small}$ explores a larger set of 10 reasoning steps to introduce more variation and enhance the dataset’s diversity. Additional dataset is generated using \genPipeline for training \PHI~\cite{abdin2024phi4technicalreport} and DeepSeek-R1-Distill-Qwen-7B~\cite{deepseekai2025deepseekr1incentivizingreasoningcapability} on their own generated dataset using \shortname (Section~\ref{sec:same_model_gen}. For \PHI-\shortname we use \PHI as $\pi$ and Phi-3-mini-4k-instruct~\cite{abdin2024phi3technicalreporthighly} as $\pi_{small}$ and for DeepSeek-R1-Qwen7B-\shortname, DeepSeek-R1-Distill-Qwen-7B as $\pi$ and DeepSeek-R1-Distill-Qwen-1.5 as $\pi_{small}$. Training hyperparameters details in Appendix~\ref{app:hyperparameters}. 


\subsection{Dataset \& Evaluation Metrics}
\textbf{Training Dataset.} We collected a large dataset of 20K math word problems with final answer ground-truth labels, primarily sampled from NuminaMath~\cite{numina_math_datasets} and MetaMath~\cite{yu2023metamath}. This dataset serves as the foundation for reasoning trajectory generation, self-correction, and preference learning.

\textbf{Evaluation Datasets. } We assess \shortname’s performance on a five different range of challenging mathematical reasoning benchmarks -- MATH-500~\cite{hendrycks2021measuringmathematicalproblemsolving}, AIME~\cite{huggingfaceAIMOaimovalidationaimeDatasets}, AMC~\cite{huggingfaceAIMOaimovalidationamcDatasets}, Olympiad Bench~\cite{he2024olympiadbench} and GSM8K~\cite{cobbe2021trainingverifierssolvemath}. Additional dataset details are in Appendix~\ref{app:experimentDetails}.

\textbf{Baselines for Comparison.} To assess the effectiveness of \shortname-trained models, we compare their performance against three categories of baselines. Frontier LLMs include state-of-the-art proprietary models such as GPT-4o~\cite{openai2024gpt4ocard}, which serve as strong upper-bound references. Open-source baselines consist of widely used models like Qwen/Qwen2.5-7B-Instruct, Qwen/Qwen2.5-1.5B-Instruct, Qwen/Qwen2.5-7B-Math-Instruct, microsoft/phi-4 and deepseek-ai/DeepSeek-R1-Distill-Qwen-7B (denoted as DeepSeek-R1-Qwen7B)
allowing for a direct comparison of \shortname’s improvements over existing open models. Additionally, we benchmark against MCTS-based models, AlphaMath~\cite{chen2024alphamathzeroprocesssupervision} and Marco-O1~\cite{zhao2024marcoo1openreasoningmodels}, both which leverage search-based techniques for mathematical reasoning. 

\textbf{Evaluation metrics}
We primarily report \textit{Pass@1 accuracy}, measuring the \text{correctness of the final answer}. Beyond this, we assess \textit{self-correction ability}, where models are prompted to review and refine their own reasoning. The evaluation process involves two stages: first, the model generates a solution step by step; then, it is prompted again to identify and fix any errors. We compare the initial accuracy to the self-corrected accuracy to quantify the model's improvement (see Appendix~\ref{app:experimentDetails} for more details). 

\section{Results \& Discussions}

\subsection{Performance Analysis (Same Model as Generation Policy)}
\label{sec:same_model_gen}

\begin{table}[!h]
\centering
\small
\resizebox{0.9\textwidth}{!}{
\begin{tabular}{|llllll|}
\hline
\multicolumn{1}{|l|}{Model}                                         & math 500      & GSM8K         & AIME          & AMC           & Olympiad      \\ \hline
\multicolumn{6}{|c|}{\textit{Baseline}}                                                                                                        \\ \hline
\multicolumn{1}{|l|}{GPT4o}                                    & 69            & 72.3          & 10            & 45.8          & 28.30         \\ \hline
\multicolumn{6}{|c|}{\textit{Same Model as Generation}}                                                                                        \\ \hline
\multicolumn{1}{|l|}{\qwenSeven}                               & 53.8          & 83.9          & 3.3           & 26.5          & 19.9          \\
\multicolumn{1}{|l|}{\qwenSeven - \shortname}                  & \textbf{63.2} & \textbf{88.3} & \textbf{7.8}  & \textbf{34.9} & \textbf{23.7} \\ \hline
\multicolumn{1}{|l|}{DeepSeek-R1-Qwen7B}              & 85.8          & 85.7          & 21.1          & 67.5          & 39.5          \\
\multicolumn{1}{|l|}{DeepSeek-R1-Qwen7B - \shortname} & \textbf{87.4} & \textbf{89.3} & \textbf{23.3} & \textbf{78.3} & \textbf{40.1} \\ \hline
\multicolumn{1}{|l|}{\PHI}                                     & 66.6          & 88.3          & 6.7           & 39.8          & \textbf{30.4}          \\
\multicolumn{1}{|l|}{\PHI - \shortname}                        & \textbf{70.0} & \textbf{93.5} & \textbf{8.9}  & \textbf{54.2} & 30.3 \\ \hline
\end{tabular}
}

\caption{Performance of \shortname (Pass@1 accuracy), with same model as generation policy. }
\label{tab:performance_same_gen}
\end{table}

Table ~\ref{tab:performance_same_gen} presents the Pass@1 accuracy comparison across mathematical reasoning benchmarks, where same model are used as the generation policy. \qwenSeven, \PHI and DeepSeek-R1-Qwen 7B are employed as the generation policy, with each model being preference-aligned respectively. 
The results demonstrate that \shortname consistently enhances performance across all evaluated models and datasets.

The \shortname-enhanced \qwenSeven achieves substantial gains with a 9.4\% increase on Math 500, 4.4\% on GSM8K, 4.5\% on AIME, and 8.4\% on AMC. Similarly, \PHI, when augmented with \shortname, exhibits comparable performance gains, achieving a 14.4\% improvement on AMC, 5.2\% on GSM8K, 3.4\% on Math 500, and 2.2\% on AIME. Additionally, DeepSeek-R1-Qwen 7B, which already demonstrates competitive performance and surpasses state-of-the-art models such as GPT-4o, also benefits from \shortname-based training. Specifically, DeepSeek-R1-Qwen 7B exhibits a 10.8\% improvement on AMC, 3.6\% on GSM8K, 2.2\% on AIME and 1.6\% on Math 500, further reinforcing the efficacy of \shortname in enhancing model performance across diverse reasoning benchmarks. 

A key advantage of \shortname lies in its versatility—its application consistently yields performance improvements across different models, regardless of their architecture. Models fine-tuned with \shortname effectively improves their reasoning capabilities, making them robust and adaptable to complex problem-solving tasks. This generalizability highlights \shortname as a model-agnostic enhancement framework that can be integrated seamlessly into various language models to improve their reasoning capabilities and overall effectiveness.

\subsection{Performance Analysis (\qwenSeven as Generation Policy)}
\label{sec:base_accuracy}

\begin{table}[!h]
\centering
\small
\resizebox{0.9\textwidth}{!}{
\begin{tabular}{|llllll|}
\hline
\multicolumn{1}{|l|}{Model}                       & math 500      & GSM8K         & AIME         & AMC           & Olympiad      \\ \hline
\multicolumn{6}{|c|}{\textit{Baselines}}                                                                                         \\ \hline
\multicolumn{1}{|l|}{GPT4o}                       & 69.8          &     72.3          & 10& 45.8          &  28.3            \\
\multicolumn{1}{|l|}{AlphaMath-7B}                & 14.8          & 33.4          & 0            & 8.4           & 8.8           \\
\multicolumn{1}{|l|}{Marco-o1}                    & 41.2          & 81            & 1.1          & 18.1          & 23.6          \\
\multicolumn{1}{|l|}{DeepSeek-R1-Qwen7B}                    & 85.8          & 85.7          & 21.1          & 67.5          & 39.5\\\hline
\multicolumn{6}{|c|}{\textit{Same Model Family as Generation}}                                                                          \\ \hline
\multicolumn{1}{|l|}{\qwenSevenMath}              & 74.4          & 92.6          & 3.3          & \textbf{45.8} & 27            \\
\multicolumn{1}{|l|}{\qwenSevenMath - \shortname} & \textbf{75.2} & \textbf{94.7} & \textbf{6.7} & 44.1          & \textbf{28.3} \\ \hline
\multicolumn{6}{|c|}{\textit{Smaller Model as Generation}}                                                                       \\ \hline
\multicolumn{1}{|l|}{\qwenOne}                    & 24.4          & 47.8          & 0            & 2.4           & \textbf{7.6}  \\
\multicolumn{1}{|l|}{\qwenOne - \shortname}       & \textbf{28.2} & \textbf{58.9} & 0            & \textbf{9.6}  & 7.3           \\ \hline
\multicolumn{6}{|c|}{\textit{Different model as Generation}}                                                                     \\ \hline
\multicolumn{1}{|l|}{\PHI}                        & 66.6          & 88.3          & 6.7          & 39.8          & \textbf{30.4} \\
\multicolumn{1}{|l|}{\PHI - \shortname}           & \textbf{69.8} & \textbf{92.1} & \textbf{6.7} & \textbf{47}   & 27.9          \\ \hline
\end{tabular}
}
\caption{Performance of \shortname on Pass@1 accuracy (\qwenSeven as the generation policy)}
\label{tab:main_result}
\end{table}

Table~\ref{tab:main_result} presents the Pass@1 accuracy, using \qwenSeven as the generation policy. It demonstrates the significant impact of \shortname in enhancing model performance. Across all tested configurations, \shortname consistently improves accuracy, particularly on challenging multi-step reasoning tasks, highlighting its effectiveness in refining model reasoning capabilities.

\textbf{Same Model Family as Generation Policy.} 
Using \qwenSeven as the generation policy, we trained \qwenSevenMath, a math-specialized variant already optimized for reasoning, \shortname further boosts performance, adding 0.8\% on Math 500, 2.1\% on GSM8K, 3.4\% on AIME, and 1.3\% on Olympiad. These results demonstrate that \shortname effectively refines model reasoning without requiring larger-scale models or external supervision.

\textbf{Smaller Model as Generation Policy.}
To evaluate \genPipeline’s robustness, we trained \qwenOne, a significantly smaller 1.5B model. Despite its constraints, \shortname dramatically enhances its reasoning abilities, improving GSM8K by 11.1\%, AMC by 7.2\%, and Math 500 by 3.8\%. This underscores \shortname's capability to extract meaningful learning signals even from limited-capacity models, making high-quality preference optimization feasible for smaller architectures.

\textbf{Different Class of Model as Generation Policy}
Beyond Qwen models, we trained \PHI, a 14B instruct-tuned model using \shortname to achieve notable gains of 3.2\% on Math 500, 3.8\% on GSM8K, and 7.2\% on AMC, reinforcing the versatility of our approach in improving diverse model architectures. These results highlight \shortname's ability to enhance reasoning across varying model families, demonstrating its broad applicability in advancing reasoning.

\textbf{Performance of AlphaMath, Marco-o1, GPT-4o, and DeepSeek-R1.} 

Among the baseline models, AlphaMath-7B, a math-specialized model integrating MCTS-based reasoning with process supervision and step-level evaluation, struggles on complex multi-step reasoning tasks. It achieves only 14.8\% on Math 500 and 8.4\% on AMC, indicating that despite structured reasoning enhancements, its effectiveness remains limited for challenging mathematical problems. Similarly, Marco-o1, an open-source model trained on diverse reasoning datasets, including Open-O1 CoT data and synthetic MCTS rollouts, scores 41.2\% on Math 500 and 18.1\% on AMC. Both models fall short of the \shortname-enhanced Qwen2.5-7B, demonstrating the superiority of self-evolution data generation in refining mathematical reasoning.

Compared to GPT-4o, \shortname-trained models achieve competitive or superior performance on key benchmarks. Notably, Qwen2.5-7B-Math-\shortname surpasses GPT-4o on Math 500 (75.2\% vs. 69.8\%) and performs on par in Olympiad (28.3\% vs. 28.3\%).  GPT-4o maintains a significant edge on AIME, highlighting areas where generalist models still outperform fine-tuned alternatives. These results underscore \shortname’s effectiveness in structured mathematical reasoning.

Finally, we evaluate DeepSeek-R1 Distilled Qwen 7B~\cite{deepseekai2025deepseekr1incentivizingreasoningcapability} using the same experimental setup and observe the highest performance gains. DeepSeek-R1 significantly outperforms GPT-4o, with an average improvement of 14.5\% across benchmarks and an exceptional 21.67\% gain on AMC (45.8\% → 67.5\%).

\begin{table}[]
\centering
\resizebox{0.9\textwidth}{!}{
\begin{tabular}{|llllll|}
\hline
\multicolumn{1}{|l|}{Model}                      & math 500       & GSM8K          & AIME          & AMC            & Olympiad       \\ \hline
\multicolumn{6}{|c|}{\textit{Baselines}}                                                                                             \\ \hline
\multicolumn{1}{|l|}{GPT4o}                      & 69.4          &  71.4              & 10.0 & 48.2          & 28.0\\
\multicolumn{1}{|l|}{AlphaMath-7B}               & 0.6           & 3.2           & 0.0          & 1.2           & 1.5           \\
\multicolumn{1}{|l|}{Marco-o1}                   & 57.2          & 53.4          & 2.2          & 21.7          & 19.1          \\
\multicolumn{1}{|l|}{DeepSeek-R1-Qwen7B}                   & 86.4& 87.6& 28.9& 77.1& 44.4\\\hline
\multicolumn{6}{|c|}{\textit{Same Model as Generation}}                                                                              \\\hline
\multicolumn{1}{|l|}{\qwenSeven}                & 63.6          & 86.6          & \textbf{6.7} & 31.3          & 22.7          \\
\multicolumn{1}{|l|}{\qwenSeven - \shortname-SC}      & \textbf{68.4 }         & \textbf{88.6}          & \textbf{6.7}          & \textbf{37.3}          & \textbf{23.3}          \\ \hline
\multicolumn{1}{|l|}{\qwenSevenMath}           & \textbf{73.4} & 87.1          & 3.3          & \textbf{51.8} & 25.8          \\
\multicolumn{1}{|l|}{\qwenSevenMath - \shortname-SC} & 71.0          & \textbf{87.7} & \textbf{5.6} & 48.2          & \textbf{26.7} \\ \hline
\multicolumn{6}{|c|}{\textit{Smaller Model as Generation}}                                                                           \\ \hline
\multicolumn{1}{|l|}{\qwenOne}              & 17.0          & 36.8          & \textbf{3.3} & 4.8           & 7.1           \\
\multicolumn{1}{|l|}{\qwenOne - \shortname-SC}    & \textbf{26.0} & \textbf{49.8} & 0.0          & \textbf{9.6}  & \textbf{9.3}  \\ \hline
\end{tabular}
}
\caption{Performance on Self-correction Accuracy (\qwenSeven as the generation policy).}
\label{tab:selfCorrection}
\end{table}

\begin{table*}[!t]
\centering
\resizebox{0.9\textwidth}{!}{

\begin{tabular}{|lllllllllll|}
\hline
\multicolumn{1}{|c|}{\multirow{2}{*}{Model}}             & \multicolumn{5}{c|}{Base Accuracy}                                                                     & \multicolumn{5}{c|}{Self Correction Accuracy}                                     \\ \cline{2-11} 
\multicolumn{1}{|c|}{}                                   & math 500       & GSM8K          & AIME          & AMC            & \multicolumn{1}{l|}{Olympiad}       & math 500       & GSM8K          & AIME          & AMC            & Olympiad       \\ \hline
\multicolumn{11}{|c|}{\textit{Same Model as Generation}}                                                                                                                                                                                                       \\ \hline
\multicolumn{1}{|l|}{\qwenSeven - Base}                 & 53.8          & 83.9          & 3.3          & 26.5          & \multicolumn{1}{l|}{19.9}          & 63.6          & 86.6          & 6.7          & 31.3          & 22.7          \\
\multicolumn{1}{|l|}{\qwenSeven - Self Generation}           & 62.2          & \textbf{88.7} & 4.4          & 26.5          & \multicolumn{1}{l|}{\textbf{23.6}} & 64.8          & \textbf{89.8} & 6.7          & 36.1          & 25.4          \\
\multicolumn{1}{|l|}{\qwenSeven - Self Correction}      & 58.4          & 85.4          & 3.3          & 33.7          & \multicolumn{1}{l|}{20.9}          & 66.0          & 87.9          & 5.6          & 37.3          & \textbf{24.8} \\
\multicolumn{1}{|l|}{\qwenSeven - \shortname}               & \textbf{63.2} & 88.3          & \textbf{7.8} & \textbf{34.9} & \multicolumn{1}{l|}{\textbf{23.7}} & \textbf{68.4} & 88.6          & \textbf{6.7} & \textbf{37.3} & 23.3          \\ \hline
\multicolumn{11}{|c|}{\textit{Math tuned Model}}                                                                                                                                                                                                               \\ \hline
\multicolumn{1}{|l|}{\qwenSevenMath}                   & 74.4          & 92.6          & 3.3          & \textbf{45.8} & \multicolumn{1}{l|}{27.0}          & 73.4          & 87.1          & 3.3          & \textbf{51.8} & 25.8          \\
\multicolumn{1}{|l|}{\qwenSevenMath - Self Generation}      & \textbf{75.4} & 93.9          & 4.4          & \textbf{47.0} & \multicolumn{1}{l|}{27.4}          & \textbf{73.8} & 88.0          & 4.4          & 47.0          & 25.2          \\
\multicolumn{1}{|l|}{\qwenSevenMath - Self Correction} & 73.2          & 93.4          & 6.7          & 44.6          & \multicolumn{1}{l|}{25.7}          & 72.6          & 88.4          & 4.4          & 48.2          & 26.6          \\
\multicolumn{1}{|l|}{\qwenSevenMath - \shortname}          & \textbf{75.2} & \textbf{94.7} & \textbf{6.7} & 44.1          & \multicolumn{1}{l|}{\textbf{28.3}} & 71.0          & \textbf{87.7} & \textbf{5.6} & 48.2          & \textbf{26.7} \\ \hline
\end{tabular}
}
\caption{Ablation study on \shortname by training model on different stages of the \genPipeline pipeline. }
\label{tab:ablation}
\end{table*}
\subsection{Self-Correction Capabilities}
\label{sec:self_correction}

We evaluate the self-correction ability of \shortname-trained models, measuring their capacity to identify and rectify mistakes in their reasoning chains. The evaluation spans \textbf{Qwen2.5-7B, Qwen2.5-1.5B, and Qwen2.5-7B-Math}, comparing their base performance with their \shortname-SelfCorrection (\shortname-SC) versions. As shown in Table~\ref{tab:selfCorrection}, \shortname-SC significantly enhances self-correction capabilities, leading to improved  accuracy.

\textbf{Improvements Across Model Variants.}
{Qwen2.5-7B-\shortname-SC} shows a {4.8-point improvement on MATH 500} (63.6\% $\rightarrow$ 68.4\%), {2.0 points on GSM8K} (86.6\% $\rightarrow$ 88.6\%), and {6.0 points on AMC} (31.3\% $\rightarrow$ 37.3\%). These gains demonstrate that \shortname-SC helps the model effectively refine its reasoning and correct errors.

{Qwen2.5-1.5B-\shortname-SC} produces even larger improvements, with a {9.0-point increase on MATH 500} (17.0\% $\rightarrow$ 26.0\%), {13.0 points on GSM8K} (36.8\% $\rightarrow$ 49.8\%), and {4.8 points on AMC} (4.8\% $\rightarrow$ 9.6\%). The greater impact on smaller models highlights the importance of self-evolution, which helps compensate for weaker reasoning.

{Qwen2.5-7B-Math-\shortname-SC} also benefits from self-correction, with gains on {GSM8K (+0.6\%)}, {AIME (+2.3\%)}, and {Olympiad (+0.9\%)}. However, performance remains stable on MATH 500 and AMC, suggesting that models optimized for mathematical reasoning still see benefits.

\textbf{Comparison with Baselines.}
\shortname-SC significantly outperforms {MCTS-based self-correction approaches} such as {AlphaMath} and {Marco-o1}. Specifically, {Qwen2.5-7B-\shortname-SC surpasses AlphaMath by 43.5\% and Marco-o1 by 14.0\% in self-correction accuracy.}

Compared to {GPT-4o}, \shortname-SC achieves {comparable or superior performance on MATH 500} (68.4\% vs. 69.4\%), {GSM8K} (88.6\% vs. 71.4\%), and {Olympiad} (23.3\% vs. 28.0\%). However, similar to the {Pass@1 accuracy trends in Section~\ref{sec:base_accuracy}}, challenges persist on AIME, where GPT-4o maintains a strong advantage.

\vspace{-10pt}
\section{Ablation Study}  
To evaluate the contribution of each stage in \shortname, we isolate \textbf{Self-Generation} (Stage 1 + relevant Stage 3) and \textbf{Self-Correction} (Stage 2 + relevant Stage 3). This analysis shows how structured reasoning generation and iterative correction independently and jointly enhance model performance.

\textbf{Impact of Self-Evolution Stages.}  
Table~\ref{tab:ablation} highlights three key findings:  

(i) \textbf{Self-Generation} improves base accuracy significantly, with \qwenSeven gaining +4.4\% and \qwenSevenMath +2.0\%. The largest improvements occur on MATH 500 (+8.4\%), GSM8K (+4.8\%), and Olympiad (+3.7\%), demonstrating that structured reasoning trajectories strengthen initial predictions.  

(ii) \textbf{Self-Correction} enhances robustness, particularly for complex multi-step problems. AMC sees the highest gain (+6.0\% for \qwenSeven), indicating that models benefit from iterative refinement when logical consistency is critical. Notably, self-correction proves most impactful in challenging datasets, where verifying and revising reasoning steps is essential.  

(iii) \textbf{Full \shortname pipeline (Self-Generation + Self-Correction)} delivers the best performance, increasing base accuracy by 6.0\% on average and surpassing both individual strategies in self-correction accuracy. This highlights the synergy between structured reasoning generation and iterative refinement in improving model reliability.  

\textbf{Math-Tuned vs. General Models.}  
For math-specific tuning (\qwenSevenMath), self-generation alone provides consistent base accuracy gains (+2.0\%), confirming the importance of structured reasoning. Self-correction, while less impactful overall, notably boosts AMC (+4.4\%), suggesting that correction mechanisms are most effective in highly structured problem types.  

By integrating both approaches, \shortname achieves the highest overall gains, particularly in AIME (+3.4\%) and Olympiad (+1.3\%), reinforcing its adaptability across diverse problem sets. These findings validate that iterative self-evolution --combining reasoning generation and error correction -- is essential for advancing multi-step mathematical reasoning.  

\vspace{-10pt}
\section{Conclusion}
We present \shortname, a self-evolving data generation framework that enhances SLMs' reasoning through iterative learning. By integrating self-generation, self-correction, and diversity induction, \shortname enables models to refine their reasoning autonomously, improving step-by-step mathematical problem-solving without extensive human-labeled data. A key innovation, {\tt Pruned MCTS}, optimizes reasoning trajectories by selectively retaining high-quality rollouts while filtering out suboptimal paths, ensuring more efficient and reliable learning. Our evaluations demonstrate substantial gains across benchmarks, highlighting self-correction, diverse reasoning, and {\tt Pruned MCTS} in improving robustness and reducing error propagation. Ablation studies confirm their essential role in handling complex problems. \shortname\ moves SLMs closer to frontier LLMs, advancing self-improving A
I for multi-step reasoning.

\section{Limitations}
While \shortname\ significantly enhances multi-step reasoning, certain aspects can be further improved. First, our approach relies on MCTS rollouts, which, despite pruning, remain computationally intensive for large-scale training. Future work could explore more efficient search strategies or adaptive pruning techniques to reduce overhead. Second, while our self-correction mechanism refines incorrect reasoning, it does not guarantee exhaustive coverage of all failure cases. Incorporating external verifiers or broader failure taxonomies could further enhance robustness. Finally, our method primarily focuses on mathematical reasoning, and its generalization to other structured domains, such as program synthesis or theorem proving, warrants further exploration. Despite these limitations, our framework provides a scalable and automated approach to preference learning, marking a significant step toward improving reasoning in LLMs.


\if
0
\section{Engines}

To produce a PDF file, pdf\LaTeX{} is strongly recommended (over original \LaTeX{} plus dvips+ps2pdf or dvipdf). Xe\LaTeX{} also produces PDF files, and is especially suitable for text in non-Latin scripts.

\section{Preamble}

The first line of the file must be
\begin{quote}
\begin{verbatim}
\documentclass[11pt]{article}
\end{verbatim}
\end{quote}

To load the style file in the review version:
\begin{quote}
\begin{verbatim}
\usepackage[review]{acl}
\end{verbatim}
\end{quote}
For the final version, omit the \verb|review| option:
\begin{quote}
\begin{verbatim}
\usepackage{acl}
\end{verbatim}
\end{quote}

To use Times Roman, put the following in the preamble:
\begin{quote}
\begin{verbatim}
\usepackage{times}
\end{verbatim}
\end{quote}
(Alternatives like txfonts or newtx are also acceptable.)

Please see the \LaTeX{} source of this document for comments on other packages that may be useful.

Set the title and author using \verb|\title| and \verb|\author|. Within the author list, format multiple authors using \verb|\and| and \verb|\And| and \verb|\AND|; please see the \LaTeX{} source for examples.

By default, the box containing the title and author names is set to the minimum of 5 cm. If you need more space, include the following in the preamble:
\begin{quote}
\begin{verbatim}
\setlength\titlebox{<dim>}
\end{verbatim}
\end{quote}
where \verb|<dim>| is replaced with a length. Do not set this length smaller than 5 cm.

\section{Document Body}

\subsection{Footnotes}

Footnotes are inserted with the \verb|\footnote| command.\footnote{This is a footnote.}

\subsection{Tables and figures}

See Table~\ref{tab:accents} for an example of a table and its caption.
\textbf{Do not override the default caption sizes.}

\begin{table}
  \centering
  \begin{tabular}{lc}
    \hline
    \textbf{Command} & \textbf{Output} \\
    \hline
    \verb|{\"a}|     & {\"a}           \\
    \verb|{\^e}|     & {\^e}           \\
    \verb|{\`i}|     & {\`i}           \\
    \verb|{\.I}|     & {\.I}           \\
    \verb|{\o}|      & {\o}            \\
    \verb|{\'u}|     & {\'u}           \\
    \verb|{\aa}|     & {\aa}           \\\hline
  \end{tabular}
  \begin{tabular}{lc}
    \hline
    \textbf{Command} & \textbf{Output} \\
    \hline
    \verb|{\c c}|    & {\c c}          \\
    \verb|{\u g}|    & {\u g}          \\
    \verb|{\l}|      & {\l}            \\
    \verb|{\~n}|     & {\~n}           \\
    \verb|{\H o}|    & {\H o}          \\
    \verb|{\v r}|    & {\v r}          \\
    \verb|{\ss}|     & {\ss}           \\
    \hline
  \end{tabular}
  \caption{Example commands for accented characters, to be used in, \emph{e.g.}, Bib\TeX{} entries.}
  \label{tab:accents}
\end{table}

As much as possible, fonts in figures should conform
to the document fonts. See Figure~\ref{fig:experiments} for an example of a figure and its caption.

Using the \verb|graphicx| package graphics files can be included within figure
environment at an appropriate point within the text.
The \verb|graphicx| package supports various optional arguments to control the
appearance of the figure.
You must include it explicitly in the \LaTeX{} preamble (after the
\verb|\documentclass| declaration and before \verb|\begin{document}|) using
\verb|\usepackage{graphicx}|.

\begin{figure}[t]
  \includegraphics[width=\columnwidth]{example-image-golden}
  \caption{A figure with a caption that runs for more than one line.
    Example image is usually available through the \texttt{mwe} package
    without even mentioning it in the preamble.}
  \label{fig:experiments}
\end{figure}

\begin{figure*}[t]
  \includegraphics[width=0.48\linewidth]{example-image-a} \hfill
  \includegraphics[width=0.48\linewidth]{example-image-b}
  \caption {A minimal working example to demonstrate how to place
    two images side-by-side.}
\end{figure*}

\subsection{Hyperlinks}

Users of older versions of \LaTeX{} may encounter the following error during compilation:
\begin{quote}
\verb|\pdfendlink| ended up in different nesting level than \verb|\pdfstartlink|.
\end{quote}
This happens when pdf\LaTeX{} is used and a citation splits across a page boundary. The best way to fix this is to upgrade \LaTeX{} to 2018-12-01 or later.

\subsection{Citations}

\begin{table*}
  \centering
  \begin{tabular}{lll}
    \hline
    \textbf{Output}           & \textbf{natbib command} & \textbf{ACL only command} \\
    \hline
    \citep{Gusfield:97}       & \verb|\citep|           &                           \\
    \citealp{Gusfield:97}     & \verb|\citealp|         &                           \\
    \citet{Gusfield:97}       & \verb|\citet|           &                           \\
    \citeyearpar{Gusfield:97} & \verb|\citeyearpar|     &                           \\
    \citeposs{Gusfield:97}    &                         & \verb|\citeposs|          \\
    \hline
  \end{tabular}
  \caption{\label{citation-guide}
    Citation commands supported by the style file.
    The style is based on the natbib package and supports all natbib citation commands.
    It also supports commands defined in previous ACL style files for compatibility.
  }
\end{table*}

Table~\ref{citation-guide} shows the syntax supported by the style files.
We encourage you to use the natbib styles.
You can use the command \verb|\citet| (cite in text) to get ``author (year)'' citations, like this citation to a paper by \citet{Gusfield:97}.
You can use the command \verb|\citep| (cite in parentheses) to get ``(author, year)'' citations \citep{Gusfield:97}.
You can use the command \verb|\citealp| (alternative cite without parentheses) to get ``author, year'' citations, which is useful for using citations within parentheses (e.g. \citealp{Gusfield:97}).

A possessive citation can be made with the command \verb|\citeposs|.
This is not a standard natbib command, so it is generally not compatible
with other style files.

\subsection{References}

\nocite{Ando2005,andrew2007scalable,rasooli-tetrault-2015}

The \LaTeX{} and Bib\TeX{} style files provided roughly follow the American Psychological Association format.
If your own bib file is named \texttt{custom.bib}, then placing the following before any appendices in your \LaTeX{} file will generate the references section for you:
\begin{quote}
\begin{verbatim}
\bibliography{custom}
\end{verbatim}
\end{quote}

You can obtain the complete ACL Anthology as a Bib\TeX{} file from \url{https://aclweb.org/anthology/anthology.bib.gz}.
To include both the Anthology and your own .bib file, use the following instead of the above.
\begin{quote}
\begin{verbatim}
% \bibliography{anthology,custom}
\end{verbatim}
\end{quote}

Please see Section~\ref{sec:bibtex} for information on preparing Bib\TeX{} files.

\subsection{Equations}

An example equation is shown below:
\begin{equation}
  \label{eq:example}
  A = \pi r^2
\end{equation}

Labels for equation numbers, sections, subsections, figures and tables
are all defined with the \verb|\label{label}| command and cross references
to them are made with the \verb|\ref{label}| command.

This an example cross-reference to Equation~\ref{eq:example}.

\subsection{Appendices}

Use \verb|\appendix| before any appendix section to switch the section numbering over to letters. See Appendix~\ref{sec:appendix} for an example.

\section{Bib\TeX{} Files}
\label{sec:bibtex}

Unicode cannot be used in Bib\TeX{} entries, and some ways of typing special characters can disrupt Bib\TeX's alphabetization. The recommended way of typing special characters is shown in Table~\ref{tab:accents}.

Please ensure that Bib\TeX{} records contain DOIs or URLs when possible, and for all the ACL materials that you reference.
Use the \verb|doi| field for DOIs and the \verb|url| field for URLs.
If a Bib\TeX{} entry has a URL or DOI field, the paper title in the references section will appear as a hyperlink to the paper, using the hyperref \LaTeX{} package.

\section*{Acknowledgments}

This document has been adapted
by Steven Bethard, Ryan Cotterell and Rui Yan
from the instructions for earlier ACL and NAACL proceedings, including those for
ACL 2019 by Douwe Kiela and Ivan Vuli\'{c},
NAACL 2019 by Stephanie Lukin and Alla Roskovskaya,
ACL 2018 by Shay Cohen, Kevin Gimpel, and Wei Lu,
NAACL 2018 by Margaret Mitchell and Stephanie Lukin,
Bib\TeX{} suggestions for (NA)ACL 2017/2018 from Jason Eisner,
ACL 2017 by Dan Gildea and Min-Yen Kan,
NAACL 2017 by Margaret Mitchell,
ACL 2012 by Maggie Li and Michael White,
ACL 2010 by Jing-Shin Chang and Philipp Koehn,
ACL 2008 by Johanna D. Moore, Simone Teufel, James Allan, and Sadaoki Furui,
ACL 2005 by Hwee Tou Ng and Kemal Oflazer,
ACL 2002 by Eugene Charniak and Dekang Lin,
and earlier ACL and EACL formats written by several people, including
John Chen, Henry S. Thompson and Donald Walker.
Additional elements were taken from the formatting instructions of the \emph{International Joint Conference on Artificial Intelligence} and the \emph{Conference on Computer Vision and Pattern Recognition}.

\fi

\bibliography{neurips_main}
\bibliographystyle{abbrvnat}
\newpage
\appendix

\section{Appendix}
\subsection{Dataset Statistics}
The dataset generation process is divided into three different stages, \((i)\) Self Generation, \((ii)\) Self Correction and \((iii)\) Diversity. To generate the dataset around 20k samples were extracte from the NuminaMath Dataset and around 10k of preference aligned pair was generated using \genPipeline. Table~\ref{tab:datasetStat} shows the complete stat of number of preference pairs generated after each stage.
\begin{table}[!h]
\centering
\begin{tabular}{|l|l|}
\hline
Total Question                    & 18644 \\ \hline
Stage I (Self Generation) - Pairs & 3238  \\
Stage II (Self Correction) - Pair & 1932  \\
Stage III (Diversity) - Pairs     & 4665  \\ \hline
Total Preference Dataset          & 9835  \\ \hline
\end{tabular}
\caption{Dataset split generated from \genPipeline}
\label{tab:datasetStat}
\end{table}

\subsection{Prompts}
\label{app:prompts}
During the generation of our dataset, we use two different prompts for Stage I (self generation) and Stage II (Self Correction) (Table~\ref{tab:prompts}).
Below are the prompts:
\begin{table*}[]
\centering
\begin{tabular}{|p{2.5cm}|p{12cm}|}
\hline
\textbf{Stage} & \textbf{Prompt} \\ \hline
Self Generation & You are a math expert. When you respond, respond only with the Solution of the final Problem, thinking step by step. At the end of the Solution, when you give your final answer, write it in the form \texttt{"Final Answer: The final answer is \textdollar answer\textdollar. I hope it is correct."} \\ \hline
Self Correction & There might be an error in the solution above because of lack of understanding of the question. Please correct the error, if any, and rewrite the solution. Only output the final solution! At the end of the Solution, when you give your final answer, write it in the form \texttt{"Final Answer: The final answer is \textdollar answer\textdollar. I hope it is correct."} \\ \hline
\end{tabular}
\caption{Prompt used for Stage I and Stage II}
\label{tab:prompts}
\end{table*}

\subsection{Hyperparameters}
\label{app:hyperparameters}
\subsubsection*{Generation - Inference}
To keep consistency we kept the same hyperparameters and the system prompts for all the DPO preference aligned models as well as the base models. temperature=0.1, top\_p=1, top\_k=50 and the prompts same as mentioned in Table~\ref{tab:prompts}. For AlphaMAth, Macro-o1, and \qwenSevenMath we used the default system prompt.

\subsubsection*{DPO Training}
For generation we trained all the models using 10 epochs, trained using bf16, and picked the checkpoint with the lowest eval loss. Other hyperparameters include, $\beta$=0.8, warmup ratio=0.2, Learning rate=1e-06. Lora rank = 64, lora\_alpha=128 and lora\_dropout=0.05.

\subsection{Experimental Details}
\label{app:experimentDetails}
\subsubsection{Evaluation Dataset Details}
\noindent 1. \textbf{MATH-500:}~\cite{hendrycks2021measuringmathematicalproblemsolving} A diverse set of high-school and competition-level math problems.\\
2. \textbf{AIME:\footnote{\url{https://huggingface.co/datasets/AI-MO/aimo-validation-aime}}} 90 problems sampled from the American Invitational Mathematics Examination (AIME 22, AIME 23, and AIME 24).\\
3. \textbf{AMC:\footnote{\url{https://huggingface.co/datasets/AI-MO/aimo-validation-amc}}} Mathematical problems from the American Mathematics Competitions.\\
4. \textbf{Olympiad Bench:}~\cite{he2024olympiadbench} A curated set of Olympiad-style math problems for assessing complex multi-step reasoning.\\
5. \textbf{GSM8K:}~\cite{cobbe2021trainingverifierssolvemath} 8.5K high quality linguistically diverse grade school math word problems.

\subsubsection{Evaluation Process}
For open-source models, all hyperparameters are kept consistent across experiments, ensuring a fair comparison. For closed-source models, default settings are used during inference. 

\subsection{MCTS Example}
\label{app:mcts_example}
\subsubsection*{Example 1}
During stage I (self generation) the geneation policy $\pi$ is provided to solve the question:

\paragraph{Question:}There were 61 parents in the program and some pupils too. The program could seat 44 people. There were 238 people present in the program. How many pupils were present in the program? 

\paragraph{Final answer:}177

\begin{figure*}[t]
  \includegraphics[width=\textwidth]{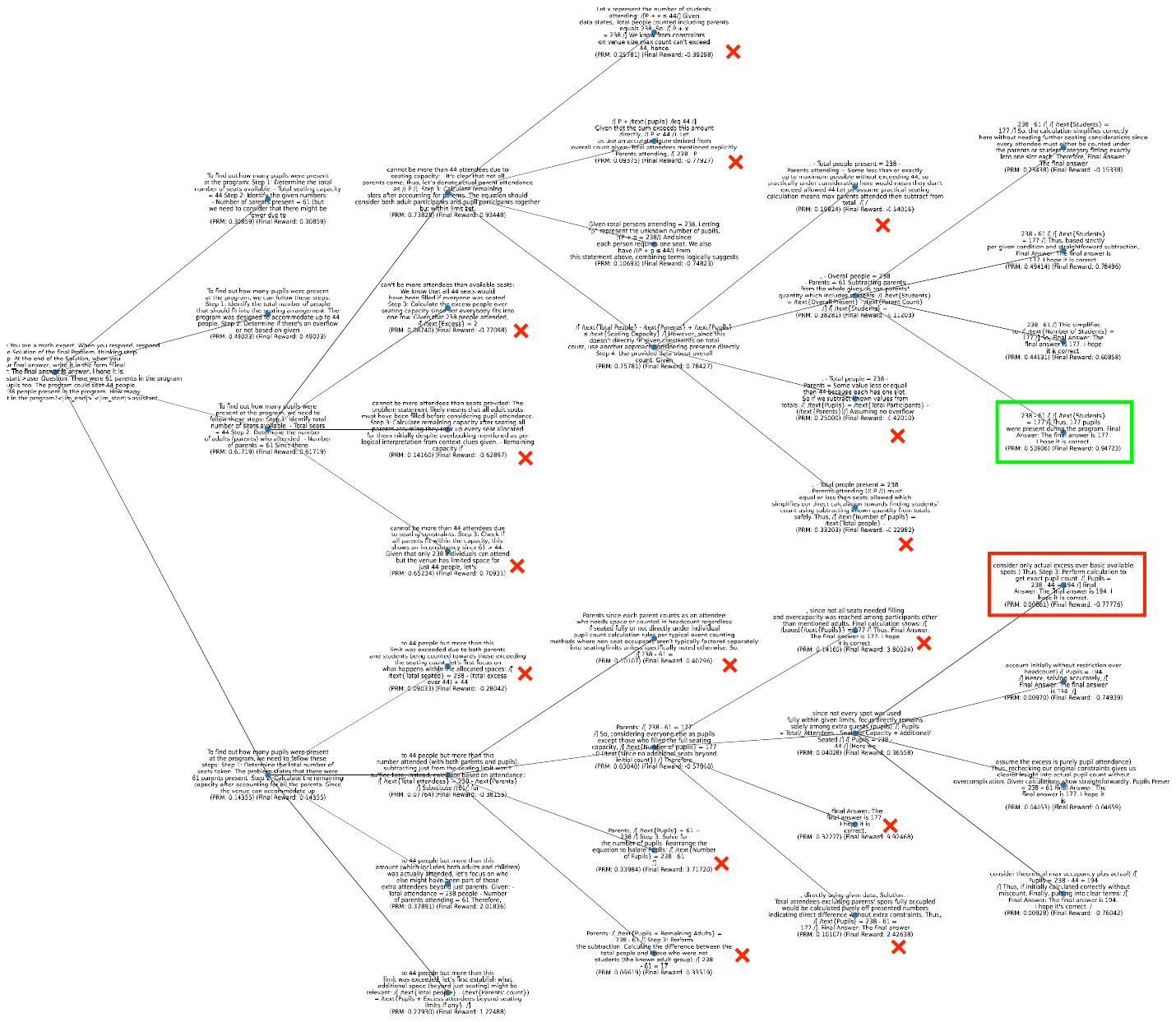}
  \caption{Stage I, rollout with question: There were 61 parents in the program and some pupils too. The program could seat 44 people. There were 238 people present in the program. How many pupils were present in the program? }
  \label{fig:example_1_stage_1}
\end{figure*}

\begin{figure*}[t]
  \includegraphics[width=\textwidth]{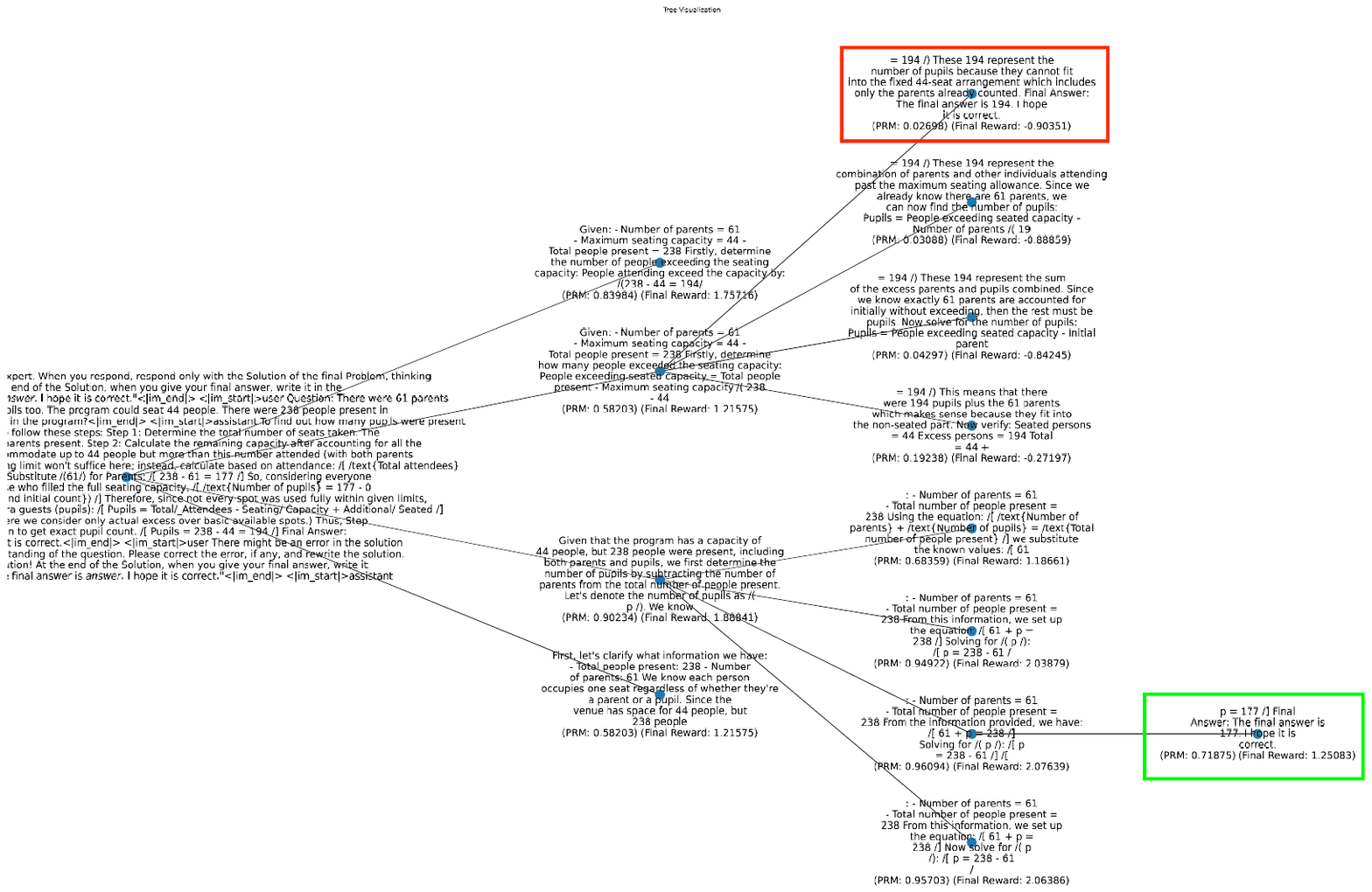}
  \caption{Stage II, rollout with question: There were 61 parents in the program and some pupils too. The program could seat 44 people. There were 238 people present in the program. How many pupils were present in the program? and incorrect solution. \\The model is prompted to identify and rectify any mistakes in the reasoning chain}
  \label{fig:example_1_stage_2}
\end{figure*}

Figure~\ref{fig:example_1_stage_1} shows the \genPipeline rollout for stage I. The positive branch reached the correct final answer and the negative branch reached the incorrect final answer, 194. Therefore the question along with the negative reasoning chain will be provided again to the generation model to self correct (Stage II, self Correction). The complete \genPipeline is showed in Figure~\ref{fig:example_1_stage_2}, where the positive branch is able to reach the correct final answer, 177.

\subsubsection{Example 2}
During stage I (self geneeration, the generation model is porvided to solve the question:

\paragraph{Question:} If the Great Pyramid of Giza is 20 feet taller than a structure that is 500 feet tall and 234 feet wider than its height, what is the total sum of its height and width in feet?

\paragraph{Final answer:}1274

\begin{figure*}[t]
  \includegraphics[width=\textwidth]{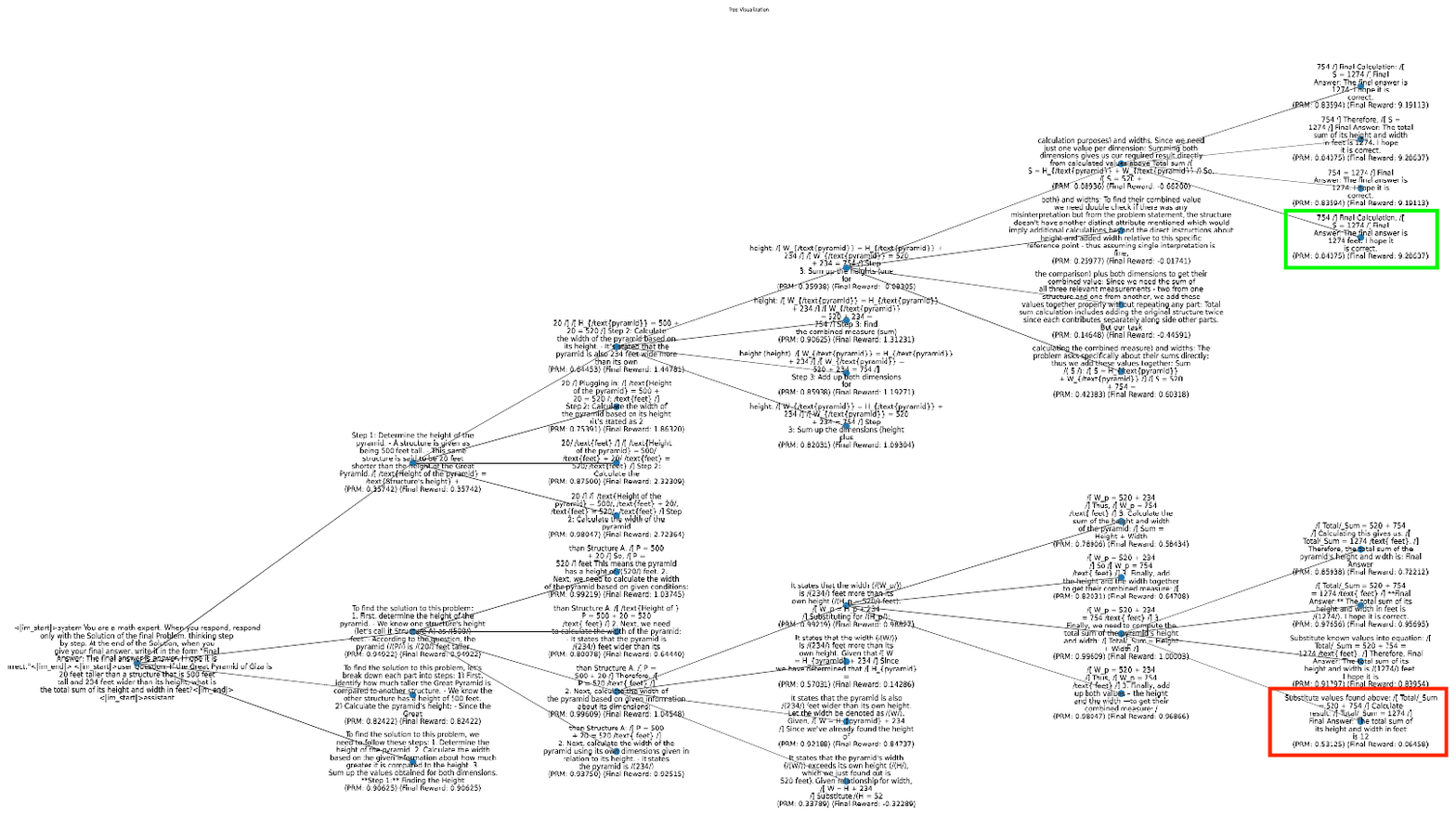}
  \caption{Stage I, rollout with question: If the Great Pyramid of Giza is 20 feet 1073 taller than a structure that is 500 feet tall and 234 1074 feet wider than its height, what is the total sum of 1075 its height and width in feet?}
  \label{fig:example_2_stage_1}
\end{figure*}

\begin{figure*}[t]
  \includegraphics[width=\textwidth]{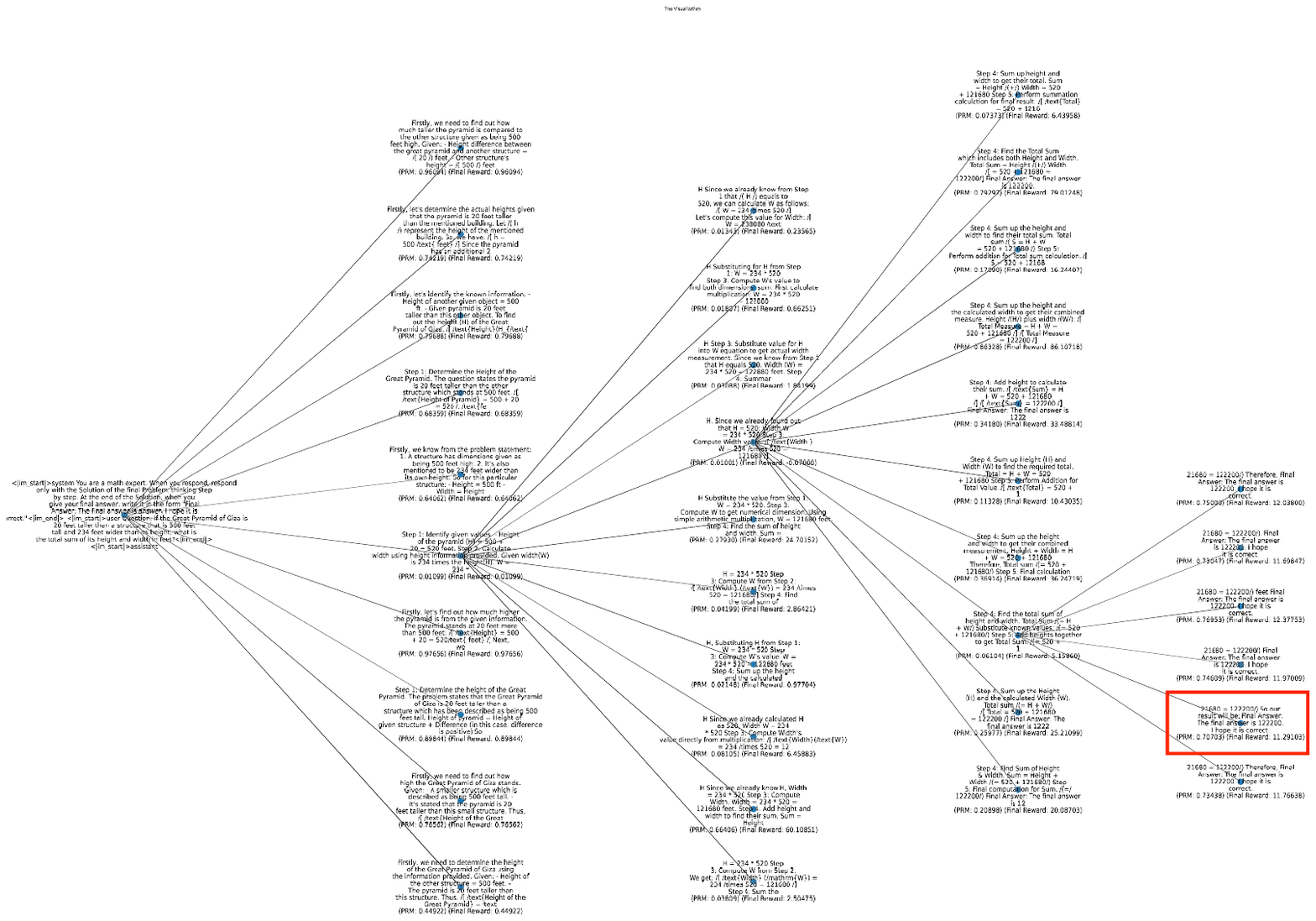}
  \caption{Stage III, rollout with question: If the Great Pyramid of Giza is 20 feet 1073 taller than a structure that is 500 feet tall and 234 1074 feet wider than its height, what is the total sum of 1075 its height and width in feet? The model is a smaller generation model and is used to generate contrastive pairs with larger exploration.}
  \label{fig:example_2_stage_3}
\end{figure*}

Figure~\ref{fig:example_2_stage_1} shows the complete \genPipeline rollout for stage I. Here both the branch are able to reach the correct final answer. Therefore the question goes to stage III to increase the diversity for the generated pairs. Figure~\ref{fig:example_2_stage_3} shows the stage III generation with a smaller generation policy and bigger rollout then Stage I. This rollout is able to successfully reach a final incorrect solution, helping achieve better preference pairs.

\label{sec:appendix}

\end{document}